# Exploring Adaptive MCTS with TD Learning in miniXCOM

**Kimiya Saadat and Richard Zhao**

Department of Computer Science, University of Calgary, Calgary, AB, Canada

{ kimiya.saadat, richard.zhao1 }@ucalgary.ca

## Abstract

In recent years, Monte Carlo tree search (MCTS) has achieved widespread adoption within the game community. Its use in conjunction with deep reinforcement learning has produced success stories in many applications. While these approaches have been implemented in various games, from simple board games to more complicated video games such as StarCraft, the use of deep neural networks requires a substantial training period. In this work, we explore on-line adaptivity in MCTS without requiring pre-training. We present MCTS-TD, an adaptive MCTS algorithm improved with temporal difference learning. We demonstrate our new approach on the game miniXCOM, a simplified version of XCOM, a popular commercial franchise consisting of several turn-based tactical games, and show how adaptivity in MCTS-TD allows for improved performances against opponents.

## Introduction

Games are suitable platforms for evaluating algorithms of artificial intelligence (AI). Games are often in environments with unambiguous rules and without external interference. They can capture the essence of real-world scenarios while maintaining a well-defined environment. AI agents that have performed well in games have been adapted to work in non-game applications (Livingston and Risse 2019). However, game playing as an AI problem can be extremely challenging due to the complexity of the game worlds. This is an interesting research area that has attracted many researchers' attention.

In recent years, Monte Carlo tree search (MCTS) has been adapted to great success in many different game applications. Its use in conjunction with deep reinforcement learning has produced success stories in many applications (Kartal et al. 2019). However, the use of deep neural networks requires substantial pre-training in general. While pre-training is possible for many applications, it is not always possible for an AI agent to have access to an environment for pre-training. Examining the ability of a game-playing AI agent to adapt to a game and its opponents on-line and during gameplay without any prior knowledge is an interesting area of research.

In this work, we present MCTS-TD, an adaptive MCTS algorithm improved with temporal difference learning. Our proposed algorithm combines both the on-line nature of MCTS and the adaptive advantages of reinforcement learning, taking the best of both worlds.

Our research question is: can MCTS be guided by reinforcement learning so that it effectively adapts to its opponents' strategies in an on-line fashion? We answer the research question by conducting experiments with MCTS-TD, comparing it to the original MCTS, another approach called SARSA-UCT, and a rule-based approach on the game miniXCOM, a turn-based grid-based tactical shooting game.

This paper makes the following contributions:

1. MCTS-TD, an improved MCTS algorithm that utilizes reinforcement learning to obtain estimated utility values of game states, and uses the utility values to guide the search in MCTS.
2. An empirical evaluation of MCTS-TD in a turn-based game to demonstrate its on-line adaptivity in different scenarios.

## Related Works

Researchers have deployed a variety of AI techniques for games. Search (Pereira et al. 2021), planning (Sauma-Chacón et al. 2020), and learning (Sieusahai et al. 2021) are all popular approaches being utilized to this day. MCTS (Coulom 2006) is a search algorithm that has become the focus of much research in gaming AI. In this section, we first explore the origin of MCTS and its variants used in creating game-playing AI. Next, reinforcement learning and temporal difference learning are introduced, alongside recent use cases. Finally, we look at notable intersections between MCTS and reinforcement learning.

## Monte Carlo Tree Search

The MCTS algorithm has shown to be widely effective in creating gaming AI agents. The basis of this algorithm includes building an asymmetric tree and searching the state space for the optimal solution while simulating the game and transferring the outcome of each episode to nodes involved in that episode.

Coulom (2006) combined Monte Carlo evaluation with tree search to create an agent that was able to play the game of GO. This resulted in the creation of the Monte Carlo tree search algorithm. Chaslot et al. (2008) proposed the use of MCTS in gaming applications. A variation of the original MCTS is the Upper Confidence bounds applied to Trees (UCT) algorithm. UCT uses UCB1 selection as the policy for selecting the next node. This algorithm is the original UCT. There are also variations of UCT such as standard UCT. Standard UCT only stores one state in the memory while original UCT stores all the visited states when memory is available and does not discount rewards in contrast to original UCT. Moreover, standard UCT treats the intermediate rewards and final rewards as the same and uses the sum of all rewards in the backpropagation step and updates all the states with the same value, while original UCT updates each state by reward and return calculated for that state (Vodopivec et al. 2017).

The MCTS algorithm and its variations have been employed in many games including Chess, Poker, and commercial video games such as StarCraft (Świechowski et al. 2022). One of the advantages of MCTS is that does not need domain-specific knowledge to perform. More recently, MCTS has been successfully deployed in the imperfect information card game Cribbage (Kelly and Churchill 2017), the board game Terraforming Mars (Gaina et al. 2021), and level generation problems (Bhaumik et al. 2020), among many other applications.

## Reinforcement Learning

In reinforcement learning, an agent learns to decide what action to take in each step while interacting with the environment and receiving rewards. The goal of reinforcement learning algorithms is to maximize the cumulative reward signal (Sutton and Barto 2018). Two important functions that are frequently used in reinforcement learning algorithms are the value function and policy function. A value function of a state can be defined as the expected cumulative reward if the agent starts from that state. A policy function is a mapping from a state to an action. The algorithms in reinforcement learning can be categorized into different categories based on certain characteristics. One categorization is based on the presence of a model of the environment. Model-based methods use a model of the environment while model-free methods do not work with a model. In model-based approaches, a model is explicitly defined with the transition probability distribution and the reward function. A model of the environment can predict the next state and the next reward using the current state and action. In model-free approaches, state values are directly learned without underlining assumptions about a model.

At the center of the model-free algorithms, there is temporal difference (TD) learning, which uses ideas from both Monte Carlo methods and dynamic programming (Sutton and Barto 2018). Temporal difference learning uses a concept named "bootstrapping" in which it can update estimates based on other estimates and it does not need to wait until the end of a game to get an outcome from the environment (Sutton and Barto 2018). One of the notable successes of temporal difference learning is TD-Gammon which was able to play backgammon at the level of the world championship (Tesauro 1995, Galway et al. 2008). In TD-Gammon, a neural network acts as an evaluation function for valid moves and is trained by the temporal difference approach.

Many of today's state-of-art reinforcement learning algorithms use temporal difference learning as part of their learning mechanism. Google's DeepMind built an agent that can play a set of Atari games. This deep reinforcement learning algorithm named Deep Q Learning can learn policies from high-dimensional input states such as images. The Q-network in this algorithm uses temporal difference as part of its learning algorithm for updating the weights (Mnih et al. 2013). Researchers have since applied reinforcement learning algorithms to many gaming applications (You et al. 2020), procedural content generation tasks (Khalifa et al. 2020), and game design and automated test challenges (Gutiérrez-Sánchez et al. 2021). While reinforcement learning approaches that utilize deep neural networks are powerful, their substantial training time is not suitable for the research question we want to explore, which is on-line adaptation.

## Intersection of MCTS and Reinforcement Learning

Researchers have examined the combined use of MCTS and reinforcement learning. One of the most prominent examples of MCTS and reinforcement learning is the AlphaGo (Silver et al. 2016) algorithm, which uses two neural networks for policy and value functions that are pre-trained with human player data as well as self-play. During on-line play, the neural networks continue to be refined by using a UCT-like algorithm. This algorithm was able to defeat the best human players in Go (Vodopivec et al. 2017). Ilhan and Etaner-Uyar (2017) proposed an approach that used MCTS rollouts as the agent's past experience and used this past experience in their reinforcement learning algorithm. They assumed that the forward model of the environment was accessible. To achieve our goal which is rapid adaptation to different strategies used by opponents, we cannot assume a fixed policy for an opponent; the next state in the environment depends on the opponent's strategy. As a result, we use the agent's past experiences to calculate TD, not the simulated rollouts as we cannot simulate accurately due to opponent's variation.

Vodopivec et al. (2017) designed an algorithm based on a temporal-difference tree search framework, SARSA-UCT($\lambda$), which used UCT as the MCTS algorithm and Sarsa as the reinforcement learning part. They were able to show that bootstrapping and TD backups were beneficial over MC backups in an MCTS-like search. In our proposed work, MCTS backups are not replaced by TD. Instead, our algorithm uses TD as a guide for the MCTS algorithm.

## Description of miniXCOM

Our game, miniXCOM, is inspired by XCOM (Take-Two Interactive 2022), a popular commercial franchise consisting of several turn-based tactical games. The first game of XCOM was released in 1994 named UFO: Enemy Unknown. In 2012, a remake of the first game was developed and released under the name XCOM: Enemy Unknown (Figure 1). In these two games, the player acts as the commander of a military squad trying to save Earth from alien invasion. These games follow a similar setup of a human squad vs. an alien squad. The player controls a squad of human soldiers on a grid-based layout, with the goal of hunting down aliens and completing objectives. The layouts of the maps vary from level to level. The two sides take turns controlling their squad, issuing commands such as move or attack. Line-of-sight is required for attacks.

Our game, miniXCOM, takes the turn-based tactical game and standardized it so that no side has an unfair disadvantage in terms of squad size or map layout. In miniXCOM, the game is played on an n-by-n grid, with a fixed and equal number of soldiers for each squad of humans and aliens. Blocks on the grid can be used as covers for these soldiers, as attacks require line-of-sight.

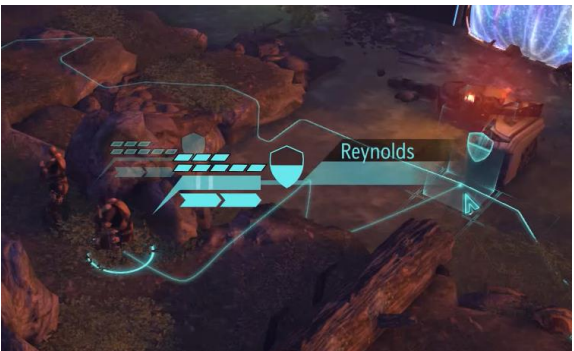

Figure 1: A grid-based battle field in the first game of the franchise, XCOM: Enemy Unknown, showing two soldiers.

### Rules of miniXCOM

Most of the rules and gameplay in miniXCOM are inspired by the original game. It is a two-player game, humans vs. aliens. While in XCOM, usually aliens are NPCs and humans are controlled by the player, in miniXCOM, either or both sides can be AI-controlled, allowing us to experiment with different AI agents.

The game is turn-based. In each turn each player has the ability to issue commands to one squad member under their control. A squad member can *move* by taking a certain maximum number of steps, *shoot* an enemy if in line-of-sight, or performing a *shoot* action immediately following a *move* action.

Moving to grid cell is only possible when there is a path to it and only if it requires at most a predetermined number of steps. A squad member can only move in four directions, up, down, left, and right, and cannot move through blocks.

Shooting a squad member is only possible if the other squad member is an enemy and there are no blocks in the line of sight. Line-of-sight is calculated by drawing a straight line between two points. Shooting immediately kills the enemy.

## Methodology

Deep learning-based approaches often require substantial training time (Gabriel et al. 2019). While these are applicable when pre-training is possible, interesting scenarios arise in applications when decisions need to be made without prior knowledge and learning or adaptation is only possible during actual gameplay. We are interested in game-playing agents that can work well in real-time without pre-training. Therefore, our work focuses on the use of MCTS-based approaches.

### MCTS

As described by Browne et al. (2012), MCTS is a tree search algorithm that repeatedly builds a game tree by playing out simulations with random moves (called rollouts) and recording the results of the different moves. Each node in the search tree accumulates information on the value of this node based on previous rollouts. This information is backpropagated through parent nodes.

In MCTS, four steps are repeated until the algorithm is stopped: selection, expansion, rollout, and backpropagation. In the selection step, a node to expand on is selected according to a selection criterion. This includes going from the root of the tree to one of its nodes that has already been added to the tree. In the expansion step, the selected node is added to the search tree. In the rollout step, the game is simulated and random actions are taken from this new node to the end of the game, and the results are totaled into a value for this node. Finally, in the backpropagation step, the cumulative score from the rollouts is backpropagated from the new node to the root. In each iteration of the four steps, one new node is added to the search tree.

As an on-line search algorithm, MCTS does not require pre-training or pre-computation of values. MCTS is also an anytime algorithm, in that it can be stopped at any time and it will return the best action found until that point based on its cumulated rollout information. This makes it an ideal candidate for our application.

## UCT

A commonly used selection criterion used in the MCTS selection step is called UCT. The UCT algorithm chooses the new node according to

$$\arg\max_{v' \in children\ of\ v} \frac{Q(v')}{N(v')} + C\sqrt{\frac{2\ln N(v)}{N(v')}} \quad (1)$$

In this formula, the first term represents exploitation while the second term represents exploration. Q(v) is the current total value at node v; N(v) is the number of times node v has been visited in the selection step; C is a constant chosen to control the rate of exploration. The two terms in the equation combine the advantage of choosing the node with the highest value with the necessity of exploring little-visited nodes.

While MCTS has many advantages in our application, it does not adapt to specific scenarios or opponent strategies. As a result, we introduce a reinforcement learning technique.

## TD Learning

In reinforcement learning, an agent is not given instructions and has to discover the appropriate actions to maximize a notion of a reward over a discrete number of steps. At time t, the agent is in state $s_t$, and takes an action $a_t$. A reward $r_t$ is received, and the agent transitions to the next state $s_{t+1}$. Actions are chosen according to a policy $\pi$.

TD learning is a model-free reinforcement learning technique that estimates the state value function under a policy. In TD learning, the values of the observed states are adjusted using the observed transitions:

$$U(s_t) = U(s_t) + \alpha(r(s_t) + \gamma U(s_{t+1}) - U(s_t)) \quad (2)$$

$U(s_t)$ is the (utility) value of state *s* at time *t*. $\alpha$ is the learning rate and $\gamma$ is the discount factor of the estimated future state value. TD learning works by adjusting the value estimates towards the ideal equilibrium that holds locally when the value estimates are correct. The equilibrium is given by the equation:

$$U(s) = r(s) + \gamma \sum_{s'} P(s'|s,\pi(s))U(s') \quad (3)$$

TD learning can be used passively to observe and learn state values without influencing the policy that determines the actions at each state. However, its learned state values can be powerful in guiding search trajectories. While there are other reinforcement learning algorithms, we chose TD learning as it is a straight-forward way of estimating state values without having access to state transition information. In the following section, we describe how we use TD learning to help provide MCTS with adaptive search trajectories.

## MCTS with TD Learning

To provide MCTS with the ability to adapt to specific opponent strategies, we propose MCTS-TD, using TD learning to provide estimates of state values at the same time as MCTS builds its search tree. While this framework can be generalized to a variety of reinforcement learning methods, we showcase the effectiveness of TD learning in our miniXCOM example and demonstrate rapid adaptation without needing pre-training. The learned state values are an effective representation of the opponent's strategy and utilizing it allows the agent to adapt to the opponent's strategy, providing an increased return.

MCTS-TD is shown in Algorithm 1. In the BestChild() function, the choosing of the next node is augmented by a new term *d U(s(v'))*, with the utility value of the child state estimated from the TD learning update. We call the *d* parameter *TD factor*, which can be adjusted to control the weight of using the utility value from TD learning.

TD learning is performed within the context of the game. The Update() function is called after MCTS-TD returns an action to the game, the action is performed, and a reward is obtained. To account for opponent actions in the game, we call the Update() function a second time, after the opponent performs an action. In this way, two updates are performed per step, and MCTS-TD takes the consequences of opponent actions into consideration.

## Experiments

We ran experiments in miniXCOM to compare MCTS-TD with the original MCTS, SARSA-UCT (Vodopivec et al. 2017), as well as a rule-based agent, RB1, described in the next section. The grid is chosen to be 6 by 6, with 2 squad members on each side and walls represented by solid blocks

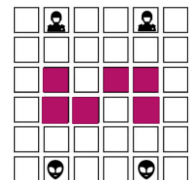

Figure 2: The setup used in the experiments.

Algorithm 1: The MCTS-TD Algorithm
**function**: MCTS-TD(state $s_0$)
    Create root node $v_0$ with state $s_0$.
    **while** within time or iteration limit **do**
        $v \leftarrow$ SelectNode($v_0$).
        $r \leftarrow$ Rollout(s(v))
        BackPropogate(v, r).
    **return** a(BestChild($v_0$, 0))

**function**: SelectNode(node v)
    **while** v is non-terminal do
        **if** v not fully expanded **then**
            choose untried action a
            v' $\leftarrow$ simulate action a at s(v)
            add v' as a new child to v
            **return** v'
        **else**
            v $\leftarrow$ BestChild(v, C)
    **return** v

**function**: Rollout(state s)
    **while** s is non-terminal **do**
        choose a from available actions in s at random
        s $\leftarrow$ simulate action a at s
    **return** reward r at s

**function**: BackPropogate(node v, reward r)
    **while** v is not null **do**
        N(v) $\leftarrow$ N(v) + 1
        Q(v) $\leftarrow$ Q(v) + r
        v $\leftarrow$ parent of v

**function**: BestChild(node v, constant C, td_factor d)

    **return** $\underset{v' \in \text{children of } v}{\arg\max} \frac{Q(v')}{N(v')} + C\sqrt{\frac{2 \ln N(v)}{N(v')}} + dU(s(v'))$

**function**: Update(td_state $s_t$, td_state $s_{t+1}$, td_reward r)
    **if** td_state s not in S **then**
        Add s to S
        U(s) = 0
    **else**
        $U(s_t) = U(s_t) + \alpha(r(s_t) + \gamma U(s_{t+1}) - U(s_t))$

as shown in Figure 2. The two sides take turn moving. In each turn, one squad member can be moved by a maximum of three grid cells, horizontally or vertically, and can shoot at an opponent squad member if there is line of sight.

The experiments consisted of comparing the results of MCTS-TD with the original MCTS, SARSA-UCT, and RB1. In each set of pair-wise experiments, 50 rounds of game were played out in one run, and this process was repeated 20 times to produce 20 runs of 50 rounds each. To avoid any potential advantages associated with being the first player or the second player, the two sides took turns making the first move – each side going first in 25 rounds.

A round ends in a draw if no side wins after making 20 moves. For MCTS-TD, TD state utility values were cleared at the start of each run, ensuring that the agent started the game with no prior knowledge from pre-training. TD learning retained its utility values within the 50 rounds to demonstrate the adaptivity of the agent. The TD factor was set to 1; learning rate was set to 0.8; the exploration constant for MCTS was to set to $1/\sqrt{2}$. These values are kept consistent across all algorithms in the experiments. We implemented the experiment agents in Python 3. The experiments were set up to run on a machine with Intel i5-9300H CPU running at 2.4 GHz.

## Results

RB1 is a rule-based agent that always picks the left corridor on the grid to launch its attack. Its rules are that if a squad member has an enemy in the line of sight, always shoot at the enemy. If there are multiple enemies in the line of sight, choose one at random. If a squad member can move to a location that has the line of sight with an enemy, then move and shoot. Otherwise, move the left-most squad member towards the left corridor.

In MCTS-TD, the TD reward is generated by examining how the action affected the current state of the game. A reward of 10 is assigned if the current action destroyed one of the enemy squad members. A reward of -10 is assigned if the current action (by the opponent) destroyed one of the squad members on the agent's side.

For the representation of states for TD learning, a 3 by 3 grid centered around the current location of the most recently moved squad member is used. This representation ensures the state space is kept at a reasonable level, which is especially important for on-line learning and adaptation.

Figure 3 shows the results of MCTS-TD vs. RB1. MCTS-TD performed much better against RB1. MCTS-TD rapidly adapted to the specific strategy used by the rules of RB1. In the first 10 rounds, the two sides were evenly matched, and MCTS-TD clearly and consistently outmatched RB1 from that point on.

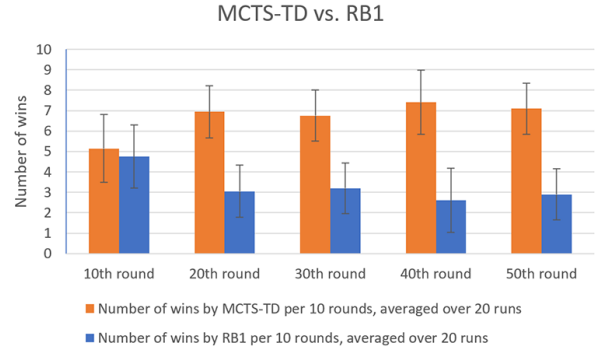

Figure 3: Results of MCTS-TD vs. RB1 averaged over 20 runs. The error bar represents one standard deviation.

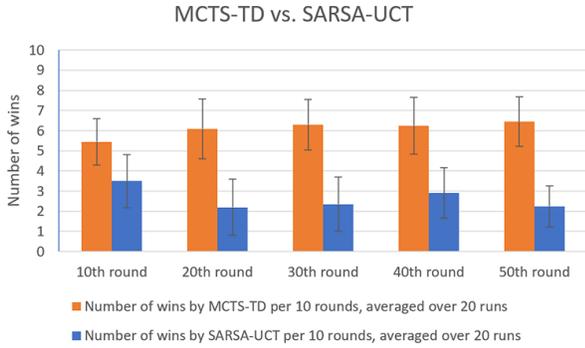

Figure 4: Results of MCTS-TD vs. SARSA-UCT averaged over 20 runs.

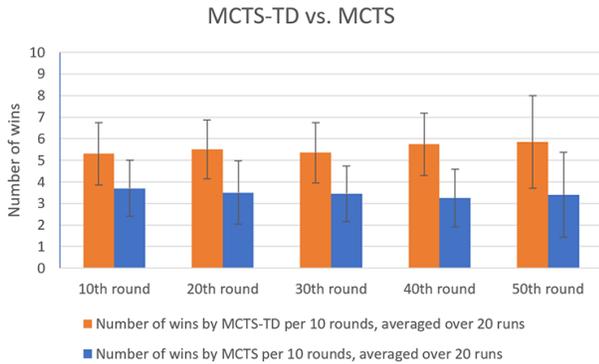

Figure 5: Results of MCTS-TD vs. MCTS averaged over 20 runs.

Figure 4 shows the results of MCTS-TD vs. SARSA-UCT. MCTS-TD clearly held an advantage over SARSA-UCT throughout the 50 rounds. On average, it was winning 6.11 rounds per 10 rounds, holding an advantage even from the first 10 rounds. SARSA-UCT did slightly worse than RB1 against MCTS-TD as the exploration of SARSA-UCT costed some games while RB1 never needed to explore.

Figure 5 shows the results of MCTS-TD vs. MCTS. We see that although the winning rates of MCTS-TD against MCTS were lower compared to it against RB1 or SARSA-UCT, it still consistently held an advantage throughout the games. While MCTS did not deploy a specific strategy and it had a considerable amount of randomness due to the exploratory nature of MCTS, MCTS-TD still quickly adapted to learn good positions for its squad members and held a higher winning rate starting from the first 10 rounds.

Table 1 provides an overview of the combined total results from all runs. Draws are not shown in the table. The results are consistent with the discussions above and MCTS-TD outmatched other agents. The difference between the winning rates for each pair is statistically significant at 99% confidence level using a paired t-test (p-values $< 0.01$ in each case).

| Winning rate out of 10 | RB1 | MCTS | SARSA-UCT | MCTS-TD |
|---|---|---|---|---|
| MCTS vs. RB1 | 5.81 (1.42) | 4.19 (1.42) | | |
| SARSA-UCT vs. RB1 | 3.55 (1.57) | | 4.59 (1.76) | |
| MCTS-TD vs. RB1 | 3.30 (1.55) | | | 6.67 (1.6) |
| MCTS-TD vs. SARSA-UCT | | | 2.64 (1.34) | 6.11 (1.33) |
| MCTS-TD vs. MCTS | | 3.46 (1.47) | | 5.55 (1.57) |

Table 1: Combined results with standard deviations in brackets. Some rounds ended in a draw due to neither side winning at time out.

## Conclusion

We are interested in examining game-playing AI agents that can perform well without opportunities for pre-training and agents that can perform on-line adaptations to take advantages of the weaknesses of opponents. In this research, we propose MCTS-TD, an adaptive MCTS algorithm with temporal difference learning. While MCTS is an effective on-line algorithm, the added power of reinforcement learning allows the algorithm to adapt to an opponent while the game is being played. We demonstrate the advantages of MCTS-TD in the game miniXCOM, a game inspired by the XCOM series of turn-based tactical games. Our results show that rapid adaptivity is promising in increasing the winning rate of the game-playing AI.

While the compressed local representation of the TD state allows for this approach to work on larger and more complex maps, we expect future work to examine other methods of representing information as part of a state, when working on more complex maps. One research direction is to examine the automated extraction of important features in a much larger state space. We will examine the effectiveness of adding reinforcement learning to MCTS in a variety of domains. MCTS-TD is applicable to other domains as there is nothing specific to the commercial-game-inspired XCOM domain in our algorithm. Our algorithm is agnostic to the TD reward that is passed in from the environment. Other domains can be explored in future work.

# References

Bhaumik, D.; Khalifa, A.; Green, M.; and Togelius, J. 2020. Tree search versus optimization approaches for map generation. In Proceedings of the AAAI Conference on Artificial Intelligence and Interactive Digital Entertainment, 24-30.

Browne, C. B.; Powley, E.; Whitehouse, D.; Lucas, S. M.; Cowling, P. I.; Rohlfshagen, P.; Tavener, S.; Perez, D.; Samothrakis, S.; and Colton, S. 2012. A survey of monte carlo tree search methods. IEEE Transactions on Computational Intelligence and AI in Games, 4(1):1–43.

Chaslot, G.; Bakkes, S.; Szita, I.; and Spronck, P. 2008. Monte-carlo tree search: A new framework for game ai. In Proceedings of the AAAI Conference on Artificial Intelligence and Interactive Digital Entertainment, 216-217.

Coulom, R. 2006. Efficient selectivity and backup operators in Monte-Carlo tree search. In Proceedings of the 5th international conference on Computers and games, 72-83.

Gabriel, V.; Du, Y.; and Taylor, M.E. 2019. Pre-training with non-expert human demonstration for deep reinforcement learning. The Knowledge Engineering Review, 34.

Gaina, R.D.; Goodman, J.; and Perez-Liebana, D. 2021. TAG: Terraforming Mars. In Proceedings of the AAAI Conference on Artificial Intelligence and Interactive Digital Entertainment, 148-155.

Galway L.; Charles D.; and Black M.; 2008. Machine learning in digital games: A survey. Artificial Intelligence Review, Springer Nature. 29(2):123-161

Gutiérrez-Sánchez, P.; Gómez-Martín, M.A.; González-Calero, P.A.; and Gómez-Martín, P.P. 2021. Reinforcement learning methods to evaluate the impact of AI changes in game design. In Proceedings of the AAAI Conference on Artificial Intelligence and Interactive Digital Entertainment, 10-17.

Ilhan, E.; and Etaner-Uyar, A.Ş. 2017. Monte Carlo tree search with temporal-difference learning for general video game playing. In Proceedings of the IEEE Conference on Computational Intelligence and Games, 317-324.

Kartal, B.; Hernandez-Leal, P.; and Taylor, M.E. 2019. Action guidance with MCTS for deep reinforcement learning. In Proceedings of the AAAI Conference on Artificial Intelligence and Interactive Digital Entertainment, 153-159.

Kelly, R.; and Churchill, D. 2017. Comparison of Monte Carlo Tree Search Methods in the Imperfect Information Card Game Cribbage. 26th Annual Newfoundland Electrical and Computer Engineering Conference.

Khalifa, A.; Bontrager, P.; Earle, S.; and Togelius, J. 2020. Pcgrl: Procedural content generation via reinforcement learning. In Proceedings of the AAAI Conference on Artificial Intelligence and Interactive Digital Entertainment, 95-101.

Livingston, S.; and Risse, M. 2019. The future impact of artificial intelligence on humans and human rights. Ethics & international affairs, 33(2): 141-158.

Mnih V.; Kavukcuoglu K.; Silver, D.; Graves, A.; Antonoglou, I.; Wierstra, D.; and Riedmiller, M.; 2013. Playing Atari with Deep Reinforcement Learning. arXiv:1312.5602.

Pereira, L.V.; Chaimowicz, L.; and Lelis, L.H.. 2021. Birds in boots: learning to play angry birds with policy-guided search. In Proceedings of the AAAI Conference on Artificial Intelligence and Interactive Digital Entertainment, 74-81.

Sauma-Chacón, P.; and Eger, M. 2020. Paindemic: A planning agent for pandemic. In Proceedings of the AAAI Conference on Artificial Intelligence and Interactive Digital Entertainment, 287-293.

Sieusahai, A.; and Guzdial, M. 2021. Explaining deep reinforcement learning agents in the Atari domain through a surrogate model. In Proceedings of the AAAI Conference on Artificial Intelligence and Interactive Digital Entertainment, 82-90.

Silver, D.; Huang, A.; Maddison, C. J.; Guez, A.; Sifre, L.; van den Driessche, G.; Schrittwieser, J.;Antonoglou, I.; Panneershelvam, V.; Lanctot, M.; Dieleman, S.; Grewe, D.; Nham, J.; Kalch-Brenner, N.; Sutskever, I.; Lillicrap, T.; Leach, M.; Kavukcuoglu, K., Graepel; T., and Hassabis, D.; 2016. Mastering the game of Go with deep neural networks and tree search. Nature, 529(7587): 484-489.

Sutton, R. S.; and Barto, A. G. 2018. Reinforcement learning: An introduction. MIT Press Ltd.

Świechowski. M.; Godlewski, K.; Sawicki, B.; and Mańdziuk J.; 2022. Monte Carlo Tree Search: A Review of Recent Modifications and Applications. arXiv:2103.04931.

Take-Two Interactive. 2022. XCOM 2. https://www.xcom.com/. Accessed: 2022-08-08.

Tesauro G. 1995. Temporal difference learning and td-gammon. Communications of the ACM, 38(3):58–68.

Vodopivec, T.; Samothrakis, S.; and Šter, B. 2017. On Monte Carlo Tree Search and Reinforcement Learning. Journal of Artificial Intelligence Research, 60:881–936.

You, Y.; Li, L.; Guo, B.; Wang, W.; and Lu, C. 2020 Combinatorial Q-Learning for Dou Di Zhu. In Proceedings of the AAAI Conference on Artificial Intelligence and Interactive Digital Entertainment, 301-307.